\documentclass[sigconf, screen,anonymous=false, nonacm]{acmart}

\AtBeginDocument{%
  \providecommand\BibTeX{{%
    \normalfont B\kern-0.5em{\scshape i\kern-0.25em b}\kern-0.8em\TeX}}}

\begin{document}

\title{Landmarks, Monuments, and Beacons: Understanding Generative Calls to Action}
\author{Victoire Hervé}
\email{contact@vherve.fr}
\orcid{0009-0009-3784-2360}
\affiliation{%
  \institution{University of Hertfordshire}
  \city{Hatfield}
  \country{United Kingdom}
}

\author{Henrik Warpefelt}
\email{research@warpefelt.se}
\orcid{0000-0002-9694-9944}
\affiliation{%
  \institution{Warpefelt Consulting LLC}
  \city{Oceanside, CA}
  \country{USA}
}

\author{Christoph Salge}
\email{c.salge@herts.ac.uk}
\orcid{0000-0001-5520-8755}
\affiliation{%
  \institution{University of Hertfordshire}
  \city{Hatfield}
  \country{United Kingdom}
}

\renewcommand{\shortauthors}{Hervé, Warpefelt and Salge.}

\begin{abstract}
Algorithmic evaluation of procedurally generated content struggles to find metrics that align with human experience, particularly for composite artefacts. Automatic decomposition as a possible solution requires concepts that meet a range of properties. To this end, drawing on Games Studies and Game AI research, we introduce the nested concepts of \textit{Landmarks}, \textit{Monuments}, and \textit{Beacons}. These concepts are based on the artefact's perceivability, evocativeness, and Call to Action, all from a player-centric perspective. These terms are generic to games and usable across genres. We argue that these entities can be found and evaluated with techniques currently used in both research and industry, opening a path towards a fully automated decomposition of PCG, and evaluation of the salient sub-components. Although the work presented here emphasises mixed-initiative PCG and compositional PCG, we believe it applies beyond those domains. With this approach, we intend to create a connection between humanities and technical game research and allow for better computational PCG evaluation.
\end{abstract}


\keywords{Theory, Procedural Content Generation, AI, Games, Human Factors}

\begin{teaserfigure}
  \includegraphics[width=\textwidth]{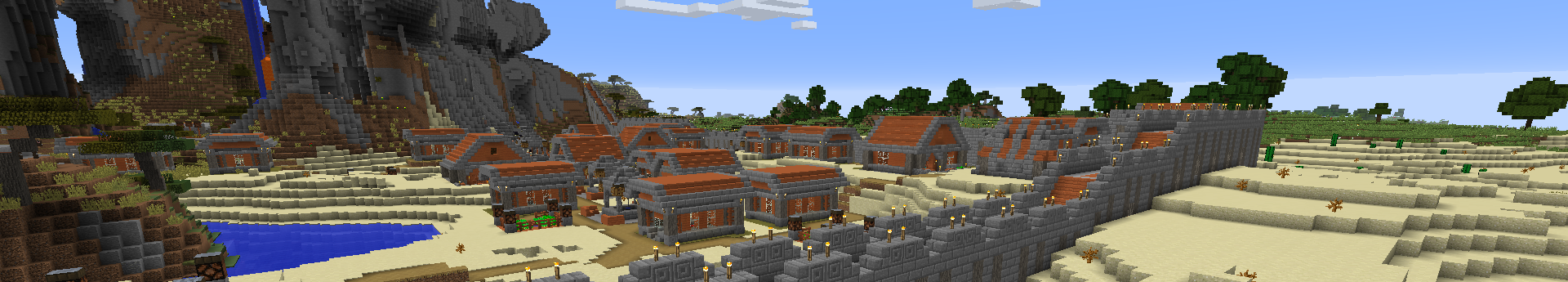}
  \caption{A Settlement in the game Minecraft.}
  \Description{}
  \label{fig:teaser}
\end{teaserfigure}

\maketitle

\section{Introduction}
This paper introduces concepts for developing metrics that can automatically evaluate complex game content in a way that aligns with human experience. Automatic game evaluation would benefit human-lead design efforts and support prototyping, but its real strength would be in the acceleration of the design-and-test loop in procedural content generation (PCG)\cite{hendrikx2013procedural} for games, such as using PCGML (machine learning)\cite{summerville2018procedural} to design levels and using automatic evaluation to adapt the content to optimise a specific desired experience. 

While there are already many existing game evaluation metrics, particularly focussed in the literature on PCG, most of them were developed to be used in the \emph{expressive range analysis} \cite{smith2010analyzing}, where they reduce games to a range of scalar characteristics with the aim of determining if two given games are different. While many of these metrics have been used ``as if’’ they capture a desired human experience - to move towards experience driven PCG\cite{yannakakis2011experience}- empirical work that looked at the alignment of common PCG metrics and player’s experience found few and only weak correlations \cite{marino2015empirical, herve2021comparing}. There are some very domain, or game, specific metrics that seem to work in limited contexts - but there exists a need for more generic metrics that can help create boundary pushing and user centric procedurally generated content. Similarly, there have been a range of proposed approaches, which often did not find their way to implementation\cite{canossa2015towards}.  

We posit that one problem that hinders the development of generic, automatic game evaluation metrics is the complexity of composite artefacts in games. By composite artefacts we mean artefacts or content that is composed or smaller components that are often similar to each other. The example that led us to this work, and also inspired some of the nomenclature later, is the design of settlements or maps in games. Settlements are made up of many houses and roads, and those are often similar to each other. Houses in turn might be made up of rooms, or furniture, again some of the components being copies of each other. There is often a hierarchical nature to these entities, and the quality of the overall artefact, the settlement is determined in part by the quality of components, and in part by how they are composed. In contrast to a simple artefacts, players might not experience all of the artefact. Instead, only a subsection of the artefact might be experienced, and that subsection is not even randomly chosen, but determined by choices and interest of the player. Experiencing some parts of the artefact might also create expectations for parts that are experienced later. For example, seeing a medieval house initially might make us think we are in a specific time period, and seeing another house from a different time period later might challenge our suspension of disbelief. This concept is in contrast to the idea of content orchestration \cite{lee2014facet}, where different types of game content, such as a level, music and character design, have to be orchestrated and aligned to work with each other, as here the content that is composed is of a similar type, and usually also has several hierarchical layers. 

We believe that the concept of a composite artefact, and by extension the introduced metrics in this paper, can be applied to composite artefacts in other game domains. For example, the soundscape of a game can be seen as a composite of sounds, or the space of items and craftables is composed of single items, or a narrative is composed of story beads, paragraphs, or even words. In all those cases you have a hierarchy, items of similar type on a hierarchy level, similarity between some components and expectation setting from one component to the next.

We believe that moving from simple to complex composite artefacts poses a strong opportunity to game evaluation metrics and their alignment to human experience. Players might focus, and have their experience determined by a small, meaningful subset of the artefact, but an automatic metric might treat all parts of the overall artefact the same. For example, when an common entropy measure was applied to the block locations in a generated Minecraft settlement, comparing the before and after maps, the measurement was mostly dominated by the large extend of the still unaltered map and its terrain\cite{herve2021comparing}. Yet the players experience was mostly affected by the few houses that were placed by the algorithm. Further studies of players self-reported reasons for judgement put spotlights on specific items, and also their compositions and relation, that were hard to capture with more traditional metrics, which were mostly count-based, or considered the configuration of the basic building block \cite{herve2023examination}s. 

To alleviate this problem, we plan to introduce a set of nested concepts that allows for an, ideally, automatic identification and decomposition into relevant sub-components, allowing for a divide and conquer approach by evaluating the sub-components, and their relationship. 
We draw upon some existing research in game studies, which we will discuss further in the related work section. Our hope here is to bridge the gap between some of the technical requirements and the conceptual ideas, by developing concepts that provide a connecting language, and have a clearly outlined route towards implementation or empirical testability, which we will outline specifically. 

The paper is structured as follows. We will first introduce the three nested concepts, Landmarks, Monument and Beacons, and provide their basic definition and properties. We will then provide a list of examples for each three types to illustrate them. Finally, we will discuss how each of those concepts would be tested, either automatically or in an empirical setting. 

\section{Related Work}
\subsection{Indicators and Actions in Video Games}

Landmarks are a concept used in numerous contexts. The most common use is in orientation: a tall buildings can be used to navigate in cities, stars in the sky point to various directions, natural elevations act as point of references in natural landscape, and so on. But similar concepts exist in other spaces. For instance, major historic events allow us to structure and navigate history.

This idea of guiding a person through an environment using large structures with a strong component of evocation has been used in entertainment, noticeably in theme parks \cite{sklar2013dream}. Visitors navigate through a large area simply by looking around them and following the trail of evocation they are most interested in. One feature of evocation is that it also creates expectations, as long as the user is able to understand them properly, meaning as intended by the designer. In a theme park, a large haunted manor is likely to contain a dark ride, which might appeal to a certain audience.

A similar concept in video games has been described by Warpefelt, who called them \textit{indicators} \cite{warpefelt2020micro}. During play time, a player observes a collection of indicators presented by the game, interprets them, and forms expectations of the game. A similar concept, focusing on narratives, are the \textit{indexes} as defined by Fernandez-Vara \cite{fernandez2011game}.

What we want to focus on in this paper is the actions provoked by landmarks. We refer to that idea as the Call to Action. We build it as an extension/specialisation of the concept of\cite{mcgrenere2000affordances}, in particular is application to games as defined by Cardona-Rivera \cite{cardona2013cognitivist}.
As stated by Cardona-Rivera, player behaviour in games cannot be understood by solely looking at the game's environment, and requires an analysis of the perceived affordances. Here in particular, we want to focus on the \textit{Perceived Affordances} \cite{cardona2013cognitivist}, i.e. the actions perceived as possible by the player.
Landmarks generate Call to Actions, that might be followed or not by a user. The same way a theme park visitor might take the action of walking toward a specific land.

All these concepts are focusing on different aspect of games, and leveraging different behaviour from the player. Our intent here is to unify all these ideas and provide a framework dedicated to PCG.


\subsection{Procedural Content Generation (PCG)}

One of the main issue of PCG is called the "Oatmeal problem" \cite{compton2016so}. Defined by Compton, the idea is that generated artefacts tend to be perceived as being too similar, no matter of how much complexity they embed. Compton also defined two types of differentiation, namely 'Perceptual Differentiation' and 'Perceptual Uniqueness'. Perceptual Differentiation 'is the feeling that this piece of content is not identical to the last', while Perceptual Uniqueness represent the idea of artefacts having its own 'character' \cite{janlert1997character}.

Both ideas of Perceptual Differentiation and Uniqueness have been investigated in the field of PCG, in particular in the research of novel ways to computationally evaluate generated content. 
One of the most common method to assess the variety of a generator is called Expressive Range Analysis \cite{smith2010analyzing}, which projects a sample of generated artefacts onto a 2D space, based on specified scalar metrics. These metrics are previously developed with the intent to easily compare two artefacts. A metric can be computed automatically, and generally addresses one characteristic of the artefact: the difficulty it introduces in the game, its look, how balanced it is, is it even playable?

One problem is these metrics are not successfully capturing what characterises an artefact. Very few empirical studies of their accuracies have been conducted, and previous results tend to show in particular their lack of embodiment \cite{summerville2017understanding, marino2015empirical, herve2021comparing}. Consequently, comparing two artefacts based on these values is often misleading, as they do not capture a significant difference. Which loops back with the challenges of capturing Perceptual Differentiation and Uniqueness.

Previous works have tried to establish new metrics with a better grounding \cite{canossa2015towards, herve2023examination, herve2022automated}, but to our knowledge, they were not reused in other practical studies. One potential explanation could be that while these new suggestions embed theories and knowledge from other field, they have not been fully translated into practical implementations that could be easy to reuse. In particular, depending on the type of game they might be used in, they might be targeting different features. 

With this paper, we aim to create a framework helping with translating such metrics into actual implementation, which target the significant features of the player experience. By doing so, we intent to support a better automated evaluation of Perceptual Differentiation and Perceptual Uniqueness.

\section{Defining Landmarks}


In this section we will introduce and define our three categories of landmarks:

\begin{itemize}
    \item Landmarks are noticeable features in the game that act as scaffolding for the gaming experience.
    \item Monuments are Landmarks with a component of evocation.
    \item Beacons are Monuments with a ``Call to Action''.
\end{itemize}

Landmarks, Monuments and Beacons assist the player in navigating the game space, both physically and cognitively. While we originally built our concepts for video game environments, we will argue they can be easily transposed to other dimension of video games, such as narrative, gameplay, music, etc. We also are interested specifically on how those can be integrated into procedurally generated content, but will describe them in the following section from a designer-agnostic perspective, and will also rely mostly on human-designed examples for illustration. 

Note that the definitions are nested, so all Monuments are also Landmarks, and all Beacons are also Monuments. Since all discussed terms are always Landmarks we will use the term Landmarks to refer to them collectively as well. 

\subsection{Landmark}

A \textit{Landmark} is a noticeable features in the game that acts as scaffolding for the gaming experience: a tower in the distance, a particular dialogue option, an item on the ground, a sound effect. For something to be a landmark it requires two properties, it has to be perceivable and stand out from the rest. Being perceivable means that the player can either experience the artefact directly through their senses (visible, audible, readable), or experience it through the game systems. But it also has to stand out from the rest of the artefacts. This can be achieved by having the landmark deviate from another similar item in a perceivable value, such as size, colour, pitch, height, etc. It might be the one giant building in the distance that is taller than the rest. There are also specific values, that are considered more noticeable than other, such as ``firefighter red’’ as a colour, or various alarm sounds. It could also be by design. For instance, in a platformer game, it can be relevant to label gaps and floating platforms as 'Landmarks' since they stand out from the ground. We will refer to this concept of standing out as 'salience,' as defined by Canossa et al. \cite{canossa2015towards}. 

The role of landmarks is to serve as anchor points to player experience, as they are often the first things players will notice, and thus be able to set expectations for the rest. They also provide a structure to the rest, by providing a sort of coordinate system that allows us to mentally relate other components to it. Like, this story beat happened after the climax, or this building is found on the other side of the tower. Yet, landmarks by itself are not necessarily memorable, often just serving their role while they are being perceived and then quickly passing as other items fill our senses. 


\subsection{Monuments}

\textit{Monuments} are \textit{Landmarks} that also have an evocative nature.
By evocation we mean they strongly relate to a previous experience of the player, and thereby conjure up a memory, a feeling, or a certain atmosphere. Monuments, in contrast to Landmarks, have a meaning that defines them beyond their strict sensory perceivability. For instance, a Monument can be a bell tower, a watch tower, a dungeon, an obelisk, etc.

Monuments are not only perceivable, they are interpreted through the lens of gameplay, narrative, or even real-life cultures and tropes \cite{herve2023examination}, i.e. they have an element of Indexical Storytelling \cite{fernandez2011game}. By evoking a specific target experience or emotion Monuments can serve two different roles. By having an emotional reaction attached to them they are more likely to be memorable and can be more easily referenced, thereby providing further structure to the content. They can also be used to guide the interpretation of the content by evoking certain target emotions to contextualise the other content. 

The evaluation and interpretation of Monuments relies on more abstract concepts and memories. The layers of interpretation might come from historical references, cultural references, personal experience, or prior in-game experiences. Some of these are still suitable for automatic evaluation, i.e. Canossa and al. \cite{canossa2015towards} suggested the use of colour theory to interpret games. Hervé and al. implemented architectural theories of isovists in Minecraft, proposing it as a predictor for the playing experience \cite{herve2022automated}. Several researchers also looked into narratives structures \cite{alvarez2022tropetwist, cardona2012indexter} and how they relate to experience. In general though, there are a range of instances where players need to have the required background, experience and memories in order to interpret the Monument correctly. While a common approach in design is to evoke experiences that are nearly universal to humans, and thereby create a reliable evocation, this can fail if what the designer considers universal experiences is not, in fact, universal.

\subsection{Beacons}

\textit{Beacons} are \textit{Monuments} that also create a `Call to Action'. By `Call to Action', we mean the content intrinsically encourages the player to perform one or several actions. It does so `intrinsically', because it leverages a players intrinsic desire to do certain actions, such as explore something interesting, without having to rely on a game-based internal reward, such as points. The actions can be directly tied to the gameplay's grammar. This concept is also closely linked to the idea of affordances, as it centres around items or entities that are designed to be interacted with in a certain way. Following from our earlier example, a massive tower, with a foreboding presence in the distance, might also encourage the player to travel there and figure out what it is. In contrast to monuments, the effect is facing forward in time, i.e. it does not connect only to a past experience and evokes it, but experiencing it affects and changes how a player will act going forward - or at least wants to act. The time-scale of this might also be quite different, from an immediate desire to get out of a current game situation, to a guiding motivation throughout the whole game, or even potentially one that affects your actions beyond the scope of the game itself. 

This design concept can take several roles. In its most immediate form it might turn a player from an observer of content to someone who interacts with the content - thereby creating a deeper, more meaningful game experience. When done well it also allows to nudge the player to do certain things, while not undermining the perception of agency. Rather than telling the player they have to do a certain thing, a beacon might make the player want to do said thing, and thus achieve the same goal, while also giving the player the impression that they could have chosen differently. This can also be used to design a game that relies fully on user-generated internal goals (which are more likely better aligned with the desired game experience of the player than designer made goals), in a game that would otherwise leave the player without direction. Finally, the idea to move the player to action beyond the scope of the game is essential if you want to design a game to affect social change or any kind of game directed at changing user behaviour, be this for the common good or economic reasons.  

This design concept relies even more on the specifics of a given player than monuments.
Things like player-types \cite{bartle1996hearts} can influence what kind of behaviour you are more inclined to pursue. Specific prior experiences or memories might also be needed to decode the `Call to Action' of a given beacon, with the most notable example being knowledge of the mechanics of a given game, i.e. an understanding of what you can do. It also means that looking at game content without any defined game, makes it hard to evaluate if a given thing would serve as a beacon. Also, while some motivations are quite universal, and can be used as hooks for designing a beacon, others are quite specific to certain groups and backgrounds of players, and thus are challenging to predict. A failure in designing a beacon could mean that players are not doing what the designer wants them to do, and thereby having an unintended game experience, or that they are feeling disoriented, and don't know what to do at all, and thus are then quitting the game.

\section{Examples of Landmarks in Games}
\subsection{Landmarks}

As stated in our definition, Landmarks are primarily noticed through senses.

In a PCG context, a good common Landmark examples are landscape features, such as tall features that can be distinguished in a distance: a large elevation or canyon for instance.
In \emph{Minecraft}, generated structure outlines (villages, temples, etc.) can be spotted from far away, even if they cannot be immediately identified. They stand out of the rest of the generated landscape, and guide the players in their exploration. 

As stated earlier, Landmarks can also stand out due to their salience, which means their distinctive characteristic with surrounding artefacts of similar kind. A sound based Landmarks could be a distinguishable sound effect, that stands out from the rest of the general soundscape. 

\begin{figure}
    \centering
    \includegraphics[width=\linewidth]{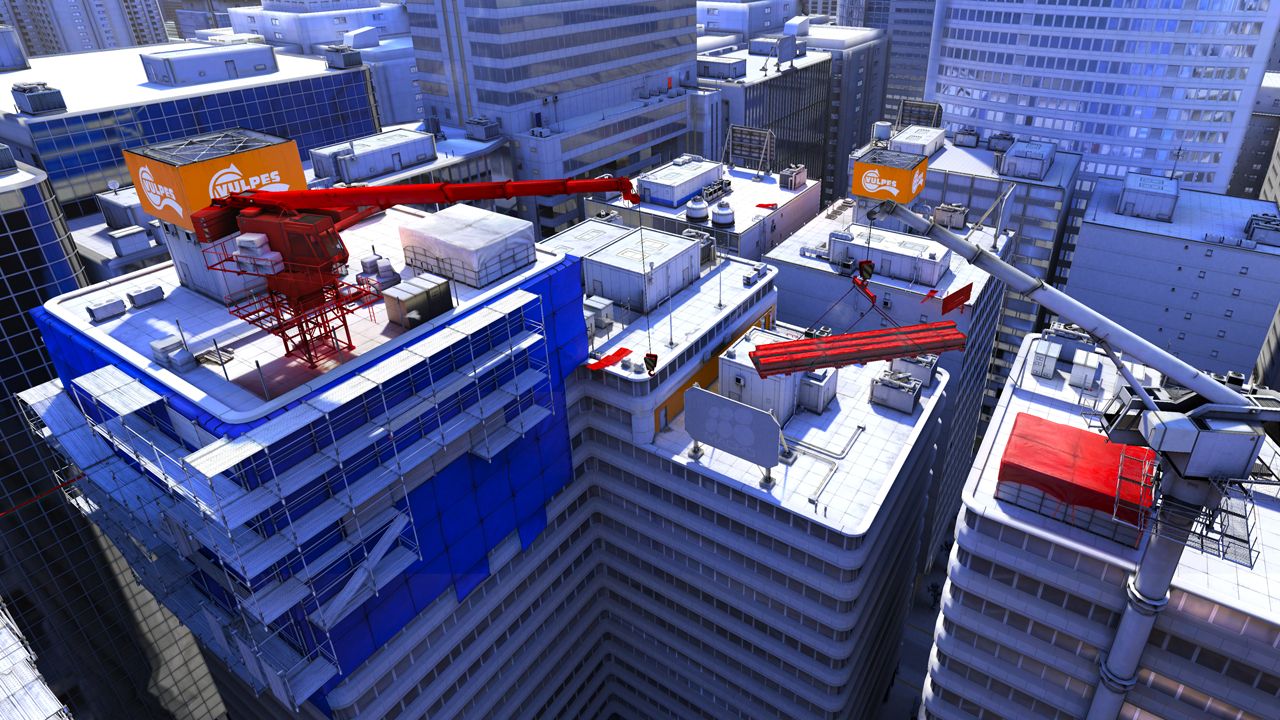}
    \caption{A \emph{Mirror's Edge} level showcase. Navigation is eased by highlighting in red part of the scenery the player can climb, or use to assist their jumps.}
    \label{fig:ME}
\end{figure}

While the reliance on player's senses is the most straightforward way, some games are designed with enhanced senses, assisting players with spotting the surrounding Landmarks.
In \emph{Mirror's Edge} \cite{game:ME}, the player is  a 'runner' which navigate through the roofs of a fictional city. Runners have developed an extra sense called "runner vision", which allows them to spot practical ways they can use to navigate the rooftops. Since the players have not developed this 'sense', useful structures are coloured in bright red, which contrasts with the overwhelming white rest of the level, as seen in Fig.\ref{fig:ME}.

Providing a list of landmarks that are just landmarks, but not monuments or beacons, from popular games is ironically quite difficult, as the lack of evocation specifically makes them less memorable, so any example that would be well known and remembered is not ``just'' a landmark, but likely a monument as well.


\subsection{Monuments}

For Monuments, we are looking at Landmarks that triggers memories, either from the current game or from past experiences, real life events or tropes.

\begin{figure}
    \centering
    \includegraphics[width=\linewidth]{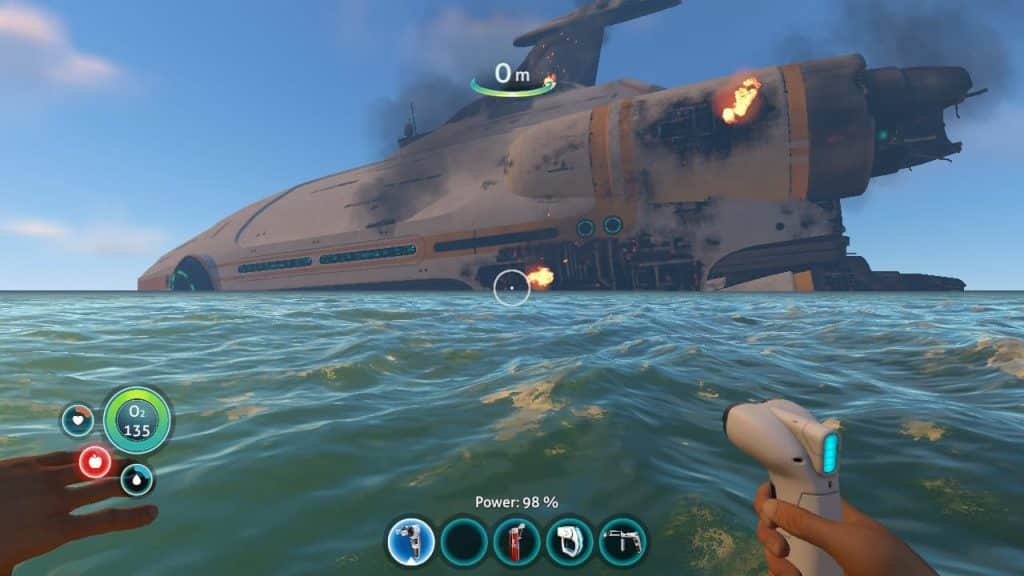}
    \caption{screenshot from the opening scenery of \emph{Subnautica}, which highlight player's crashed space-ship. It stands out of the rest of the scenery (an endless ocean), and provide context regarding the story: the player is stranded on this planet, they come from the outer space, and have access to advanced technology.}
    \label{fig:Subnotica}
\end{figure}

In \emph{Subnautica} \cite{game:Subnautica}, players start their adventure right next to a massive space-ship wreck, Fig.\ref{fig:Subnotica}. This massive structure is acting as a narrative Monument: players can easily guess they are a survivor of the crash, and are now stranded on this planet. Later in the game, when looking for some advanced items, the Monument helps with locating them, since the ship is the only trace of civilisation in the surroundings.

The planets generated in \emph{No Man's Sky} \cite{nms}, come with numerous structures. However, their look depends on their type. Obviously, crashed spaceships and trade outposts are very different, and players can identify them from afar. But generated base can have different looks based on their in-game narrative. They can be inhabited, in which case settlers will roam around them; abandoned, with some broken structures, or even invaded by aliens which can be noticed by the presence of green sludge in some areas. The procedurally generated structures can therefore convey different narratives based on some of their visual features.   

While our definition of Landmarks is a concept inspired from spatial navigation, it can be extended beyond 3D objects in a landscape, and be applied to any noticeable gameplay elements or content such as sounds, snippets of speech, etc.

A common non-visual Monuments are musical themes, that can be associated with characters or areas. Reusing these themes is a common way to indicate to the players they are about to encounter a said character, or getting close of a specific area.

\begin{figure}
    \centering
    \includegraphics[width=0.5\linewidth]{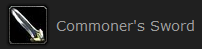}
    \caption{Commoner's Sword, a low level weapon from \emph{World of Warcraft}.}
    \label{fig:Commoner's Sword}
\end{figure}

\begin{figure}
    \centering
    \includegraphics[width=1\linewidth]{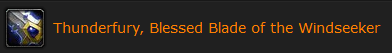}
    \caption{Thunderfury, Blessed Blade of the Windseeker, one of the rarest sword from \emph{World of Warcraft}.}
    \label{fig:thunderfury}
\end{figure}

We can also have a look at how in-game items use evocation to provide informations to the players. 
For instance, the look of a weapon can tell if it is whether a mace or a sword, but also if it is tied to certain in-game tropes, its expected rarity or quality. Same can be said for naming convention of the items, and other of its features such as attached SFX, VFX, and so on.
As a practical example of items, we will examine two weapons from \emph{World of Warcraft}(a.k.a. \emph{WoW} \cite{game:WoW}). Fig.\ref{fig:Commoner's Sword} is the in-game description of a basic sword that can be found in low level areas, and as a low value in terms of power and resale. On the opposite, Fig.\ref{fig:thunderfury} is one of the rarest and difficult sword to obtain in the entire original game. While these exacts informations might not be accessible just by looking at the icons, they however give some cues. The icon and naming of Fig.\ref{fig:Commoner's Sword} imply that it is somewhat mundane and disposable; while Fig.\ref{fig:thunderfury} as a certain grabbiness, and create expectations regarding its value. Besides, a seasoned player will interpret the text's colour of both items, gray being associated with low value items, while orange is the colour of 'legendary' items, which are among the rarest and most potent items of the game. 

\subsection{Beacons}

As a common practical example of Beacons in game, we can go back to our original approach based on visual navigation.

In the \emph{Assassin's Creed} series (from \emph{Ubisoft}), players explore historical cities, while having the ability to climb any building. The tallest buildings in the game are usually Monuments, as they are tied to a given time and place: a watchtower, a minaret, a cathedral, a bell-tower, etc. But, because of the game mechanics, they also have an intrinsic \emph{Call to Actions}: by climbing them, players can have access to some of the best vantage points in the game. This can be useful for players to plan ahead their route, locate a particular objective, or simply enjoy the view.
Some of them are even key point of the gameplay loop: they unlock a section of the map to the player. These are usually indicated in the UI, but can be spotted in game by a eagle flying around them. Therefore, every tall building in an \emph{Assassin's Creed} game act as a Beacon, with \emph{Call to Actions} to be climbed, or for some player to scrutinise them in search of an eagle.

As a PCG example, we can name generated structures in \emph{Minecraft} \cite{game:Minecraft}: villages, temples, etc. All of them have a different use in the game loop: villages have traders, temples contain treasure, etc. They can be spotted in the distance, and emit a \emph{Call to Actions} to the players: should they visit them or not.

In \emph{Skyrim} \cite{Skyrim}, a rare plant called \emph{Ninroot} emits a specific sound, which stands out from all others in the game and which players with a bit of experience can identify. This is an example of audio \emph{Call to Actions}, in which the players to switch their immediate objective, and are suggested to look for the plant and pick it up.

Another audio example, from \emph{Death Stranding} \cite{game:DeathStranding}, is the baby (named 'BB') carried by the main character. In the narrative, BB is able to perceive things beyond human senses. Whenever BB spots anything particular, they start cooing. By default, these sounds came out of the controller's speakers, putting them on a completely different audio track, and making them highly noticeable. BB's sounds act as notification, warning the player to the presence of enemies, indicating hidden locations, or even commenting their gameplay. Theses sounds once again act as Beacons, suggesting changes in the current gameplay of the players, such as dealing with the enemies, looking for resources nearby, or be more careful in general.

Finally in \emph{The Witcher 3} \cite{game:TW3}, players switch to a different view, called "Witcher senses", which highlight some Landmarks their regular human sense can hardly notice. In this mode, sounds and smells are visible, while tracks on the ground are outlined in a glowy colour. All these clues are highly noticeable. They all have different evocations, especially for players familiar with the game, and lore, as these cues can point to human, animals and even monsters. Based on their interpretations, players are subject to \emph{Call to Actions}, which then triggers numerous decisions, such as should they track the creature or not, which weapon they should use, and what strategy they should adopt.

\subsection{Discussion of Definitions}

While we defined three distinct categories, we should note two caveats. First, the definitional properties of our categories exist on a spectrum, i.e. some things are ``more'' noticeable than others, and consequently, there is no strict boundary between the categories. Using a concrete measurement, one could define a threshold, but we have to acknowledge that the question of when exactly a regular artefact turns into a landmark is somewhat subjective. This also brings up the second caveat, namely that particularly the definitions of monuments and beacons depend heavily on subjective experience, and thus can vary between different players. What is a landmark for one player might be a beacon for another player. 

\begin{figure}
    \centering
    \includegraphics[width=0.5\linewidth]{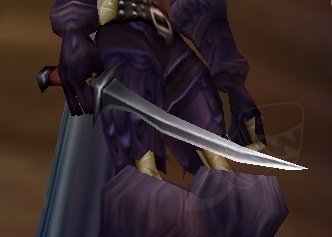}
    \caption{A \emph{World of Warcraft} character equipped with a Commoner's Sword (source WoWHead).}
    \label{fig:CommonerCharacter}
\end{figure}

\begin{figure}
    \centering
    \includegraphics[width=1\linewidth]{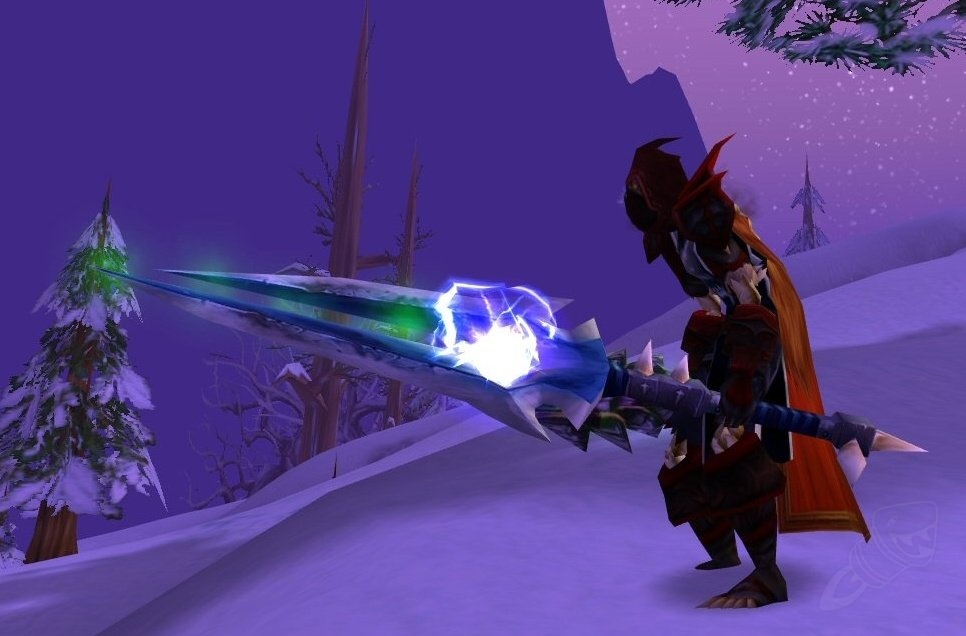}
    \caption{A \emph{World of Warcraft} character equipped with a Thunderfury (source WoWHead). We can see how the weapon has it's own specific look and makes its holder stand out.}
    \label{fig:TFCharacter}
\end{figure}

For example, a WoW player holding Thunderfury will be easily noticed due to the size and look of the weapon (Fig. \ref{fig:TFCharacter}), therefore making it a Landmark. It stands out from common, low level weapons (Fig. \ref{fig:CommonerCharacter}), which is evident when comparing the depiction of the weapons seen in Fig. \ref{fig:TFCharacter} and Fig. \ref{fig:CommonerCharacter}, even to an observer unfamiliar with WoW. But for a knowledgable player it will also be a Monument, as they can gauge the power of the weapon and even the skills of its holder, knowing what they had to accomplish to get it. Finally, in certain scenarios, such as in a player-versus-player environment, the sword might also act as a beacon, either warning players away, or causing them to prioritise the wielder. 

\subsection{Landmarks compared to regular artefacts}

While Landmarks may be the peaks in the playing experience, not every single artefact has to be a Landmark. Minor content serves to build expectation, and interpretation to the whole. As Hervé and al. suggested \cite{herve2023examination}, it is worth looking at PCG artefact as something granular, composed of multiple entities, rather than a fully holistic approach. There is something quite organic in how content builds expectation and Landmarks serve as a payoff. 

We even argue that all content in a video game is Calling to Action. What is interesting is how they do it to a different extend.
Using Minecraft as an example, you can theoretically mine any block. Depending on the context you might want to mine Cobblestone, a very common and generic block that is used mostly in early game. But you are more likely to mine iron, a resource that has it use through the entire play-through and is a bit less common. And you will certainly mine any diamonds you might encounter, one of the rarest blocks, used in the creation of some of the best items in the game.

All of these 3 blocks are emitting a Call to Action, but to very different extend. During a typical play-through, you might ignore ten of thousands of Cobblestones. Iron might make you do a detour, but depending on the situation you also might not want to deal with monsters and lava to get it. Diamond is always a go. 
1 action, 3 different calls.

With all of this taken into consideration, it is worth looking at the relation Landmarks have with more generic content. If our 1000 bowls of oatmeals were all technically unique, would they all still be engaging individually? What if instead of looking at all Commoner's swords (Fig.\ref{fig:Commoner's Sword}) as single artefacts, we otherwise look at all of them as a single entity, and add its rarity as a comparison factor with Thunderfury (Fig. \ref{fig:thunderfury})? A Commoner's Sword is also conveying its own narratives and Call to Actions.

\section{Practical implementation}

In this section we will discuss a potential implementation of a computational detection of landmarks. Starting from Landmarks, we will then discuss a noticeable artefact could be furthermore measured, in order to establish if it is Monument, or even a Beacon.  

\subsection{Landmarks}
In order to identify Landmarks, we need to compute the noticeability of the artefact. 
Some artefacts are noticeable by nature, which means they are designed to be noticed through senses (like doors in a room, fire extinguishers, sounds, mountains in the distance, etc.). In this case, we simply need to check the situation in which it becomes noticeable: making a visibility test, checking if a dialogue option is available, or if a SFX is audible, are all common game development operations. It could reuse existing metrics, as long as they are made subjective to the player's perspective. Such approach is actually similar to existing work \cite{khaleque2024evaluating, herve2022automated}, which could be tuned for Landmarks detection.
But some other artefacts, in particular gameplay elements, are too abstract to be tested through sensory means. Fortunately, the field of PCG has also already produced several metrics that aimed at capturing some of these characteristics \cite{cardona2014foreseeing, canossa2015towards}.
We would also have to check their salience \cite{canossa2015towards}, to make sure that in addition of being noticeable, the artefact also stands out from its surroundings.

Not only are we interested in checking 'if' the artefact is standing out, but also by 'how much'. This could be measured by locating in it in the generator's \emph{Expressive Range} \cite{smith2010analyzing}, and comparing it with other clusters of artefacts. However, ERA is usually applied on whole artefacts, while here it should be generated using components of a similar nature from a composite artefact. A similar approach to ERA that works on comparable components of an artefact, such as described by Herve et al. \cite{herve2023exploring}, could be used. It should be done by using metrics that can be perceived by the players: a significant height difference, an unusual colour or pitch, a twist in narrative or gameplay structures, etc.

While a generic implementation sounds conceivable, using solely common computational metrics, a per-game implementation is also sensible. 

\subsection{Monuments}

Once noticeable features are detected, we want to identify the Monuments, and therefore, the evocativeness of Landmarks. A simple test could be: "Is the artefact bringing up any memories?". Such task is difficult to automatise with the state of the art at the time of writing. However, experiments using general interpretation of games components \cite{cardona2012indexter, fendt2012achieving, alvarez2022tropetwist, herve2023examination}, already suggest this area could be investigated further. Approaching this empirical detection of Monuments would be possible. Either by asking players, during play, or shortly afterwards, how they feel or what they think about certain Landmarks. It could also be achieved by recording player's reaction during game time. Furthermore, the idea that monuments should be memorable could be used in a test focussed on recall, i.e. measuring which game elements a player can recall after a session, or which they remember in a follow-up session. Other recent AI advances, such as LLMS, might provide useful to evaluate text or images in regards to what memories or associations they conjure up in a way that could be measured or automated.

Nonetheless, interpretation is highly related to the cultural background of the player, their expectations, and what they know of the game so far. Hopefully, in the context of a video game development process, the expected interpretation could be defined in advance by the designers, and turned into an system of rules used for a dedicated implementation.
Also, the use of techniques such as our thematic analysis \cite{herve2023examination}, could help with establishing other general understanding of the mechanics of player's interpretation. 

Evocations are likely to be game specifics. However, some genres share common tropes. Tailored implementations of automated Monument identification seems required at this stage. But thinking of a common ground for custom implementation is reasonable, given that some research already aimed at such result \cite{herve2022automated, alvarez2022tropetwist, cardona2014foreseeing}.

\subsection{Beacons}

A similar reasoning and approach as for evocations can be applied to \emph{Call to Actions}. Trying to capture how a player might react or not to different \emph{Call to Actions} is quite challenging, once again because of its subjective components. Depending on the player's background, their past gaming experience, including in the current game, their tendency to react to a \emph{Call to Actions} will vary. 
Nonetheless, there is existing work on predicting player behaviour, through agent-based evaluation \cite{Baumgarten2010}, and player modelling \cite{yannakakis2013player}. Player modelling is actually one of the pillars of experience-driven PCG \cite{yannakakis2011experience}, a field from which we could draw several concepts for Beacons evaluation. However, it should be made sure that such approach revolves around player's perception. It should model player's reaction to Beacons that are noticed only, and on the intrinsic motivation coming from the \emph{Call to Actions}, rather than maximising game-based internal reward system.
Another approach for modelling player behaviour could be using player trace \cite{valls2015exploring}. Traces are obtained through recording actual gameplay sessions, which requires more resources. But they have the advantage of capturing all player's behaviour, including they reaction to new Beacons, as their in game camera moves for instance.

We suggest analysing \emph{Call to Actions} at various scale, according to our observation on \emph{composite artefacts}. For instance, in a platformer game, an enemy alone does not call for the same action that a enemy followed by a gap. Or in a RPG, an item found early in the game could have repercussion on decision made tens of hours later.
It should be kept in mind that a \emph{Call to Actions} can be part of the player's path long after being noticed, at a time when the context is more suitable for the player.

\section{Future work}

Following up to this current paper, we want to use the concepts developed in an actual implementation of automated Landmark detection and running actual empirical testing with it. We are hoping to highlight players behaviours when they interact with namely Landmarks, Monuments, and Beacons. 

Beyond practical work, there is a wide range of potential research to conduct. First, now that we have a theoretical framework for applying theoretical game studies to the field of PCG, we would have to turn these theories in actual implementations. 
The concept of Landmarks in itself raises questions, such as their optimal distribution. As an example, we could imagine using landmark's detection as a predicator for player’s fatigue.

Also, even if in our definition Landmarks can be applied to any facet of video games, there is one aspect that we haven’t addressed so far, which is the social aspect of games. Groups of players and online community have created their own Landmarks, consciously or not, and the same goes for online games. There is a great potential for researches here, rooting in player’s studies, design of multiplayer experiences (both synchronous and asynchronous), and results in this area could be a valuable addition for the video game industry.

\section{Conclusion}

In this paper, we have introduced the concept of Landmarks, Monuments, Beacons, and Call to Action. 
We have discussed how they could address several challenges in the field of PCG, regardless of the game genre or the facets. All these concepts help to better define how video game content is structured, in particular \emph{composite artefacts} and complex artefacts. With this addition, we provide a framework for PCG evaluation which focus on content more likely to be noticed, and therefore to have a larger impact on the player experience. We also suggested an actual implementation model for spotting Landmarks, underlining existing technical and theoretical work that could support such implementation. 
Furthermore, Landmarks is also a framework of communication between the fields of game's AI and game's studies. Using Landmarks, Monuments and Beacons, concepts and theories from other field could be embed in future automated evaluation techniques, offering new perspectives for the field. Additionally, during the design of a generator, it will now be possible for designers and programmers to use these concepts to communicate notions of how the generator should be designed. If a generator is able to acknowledge that it is currently producing a Landmark, it could technically take it into account, and the rest of the content could be made accordingly.
Similarly for techniques such as PCG ML \cite{summerville2018procedural}, data sets could now use Landmark based labels. In search-based PCG \cite{Togelius2011}, fitness functions could also rely on them.
Finally, by having the right focus in evaluating artefacts, it will be easier to identify what characterise them, leading to more embodied artefacts comparison. By establishing if an artefact is a Landmark, or if its Landmark components are characterising it in a certain way, we could therefore approximate its uniqueness. 

With the concepts of Landmark, \emph{composite artefacts}, and Call to Action, we hope to address several issues of the field of PCG, regarding variety and automated content evaluation. 

\bibliographystyle{ACM-Reference-Format}
\bibliography{sample-base}


\begin{thebibliography}{37}


\ifx \showCODEN    \undefined \def \showCODEN     #1{\unskip}     \fi
\ifx \showDOI      \undefined \def \showDOI       #1{#1}\fi
\ifx \showISBNx    \undefined \def \showISBNx     #1{\unskip}     \fi
\ifx \showISBNxiii \undefined \def \showISBNxiii  #1{\unskip}     \fi
\ifx \showISSN     \undefined \def \showISSN      #1{\unskip}     \fi
\ifx \showLCCN     \undefined \def \showLCCN      #1{\unskip}     \fi
\ifx \shownote     \undefined \def \shownote      #1{#1}          \fi
\ifx \showarticletitle \undefined \def \showarticletitle #1{#1}   \fi
\ifx \showURL      \undefined \def \showURL       {\relax}        \fi
\providecommand\bibfield[2]{#2}
\providecommand\bibinfo[2]{#2}
\providecommand\natexlab[1]{#1}
\providecommand\showeprint[2][]{arXiv:#2}

\bibitem[Alvarez and Font(2022)]%
        {alvarez2022tropetwist}
\bibfield{author}{\bibinfo{person}{Alberto Alvarez} {and} \bibinfo{person}{Jose Font}.} \bibinfo{year}{2022}\natexlab{}.
\newblock \showarticletitle{TropeTwist: Trope-based Narrative Structure Generation}. In \bibinfo{booktitle}{\emph{Proceedings of the 17th International Conference on the Foundations of Digital Games}}. \bibinfo{pages}{1--8}.
\newblock


\bibitem[Bartle(1996)]%
        {bartle1996hearts}
\bibfield{author}{\bibinfo{person}{Richard Bartle}.} \bibinfo{year}{1996}\natexlab{}.
\newblock \showarticletitle{Hearts, clubs, diamonds, spades: Players who suit MUDs}.
\newblock \bibinfo{journal}{\emph{Journal of MUD research}} \bibinfo{volume}{1}, \bibinfo{number}{1} (\bibinfo{year}{1996}), \bibinfo{pages}{19}.
\newblock


\bibitem[{Bethesda Game Studio}(2011)]%
        {Skyrim}
\bibfield{author}{\bibinfo{person}{{Bethesda Game Studio}}.} \bibinfo{year}{2011}\natexlab{}.
\newblock \bibinfo{title}{The Elder Scrolls V: Skyrim}.
\newblock
\newblock
\newblock
\shownote{Video game}.


\bibitem[{Blizzard Entertainment}(2004)]%
        {game:WoW}
\bibfield{author}{\bibinfo{person}{{Blizzard Entertainment}}.} \bibinfo{year}{2004}\natexlab{}.
\newblock \bibinfo{title}{World of Warcraft}.
\newblock
\newblock
\newblock
\shownote{Video game}.


\bibitem[Canossa and Smith(2015)]%
        {canossa2015towards}
\bibfield{author}{\bibinfo{person}{Alessandro Canossa} {and} \bibinfo{person}{Gillian Smith}.} \bibinfo{year}{2015}\natexlab{}.
\newblock \showarticletitle{Towards a procedural evaluation technique: Metrics for level design}. In \bibinfo{booktitle}{\emph{The 10th International Conference on the Foundations of Digital Games}}. sn, \bibinfo{pages}{8}.
\newblock


\bibitem[Cardona-Rivera et~al\mbox{.}(2014)]%
        {cardona2014foreseeing}
\bibfield{author}{\bibinfo{person}{Rogelio Cardona-Rivera}, \bibinfo{person}{Justus Robertson}, \bibinfo{person}{Stephen Ware}, \bibinfo{person}{Brent Harrison}, \bibinfo{person}{David Roberts}, {and} \bibinfo{person}{R Young}.} \bibinfo{year}{2014}\natexlab{}.
\newblock \showarticletitle{Foreseeing meaningful choices}. In \bibinfo{booktitle}{\emph{Proceedings of the AAAI Conference on Artificial Intelligence and Interactive Digital Entertainment}}, Vol.~\bibinfo{volume}{10}. \bibinfo{pages}{9--15}.
\newblock


\bibitem[Cardona-Rivera et~al\mbox{.}({[n.\,d.]})]%
        {cardona2012indexter}
\bibfield{author}{\bibinfo{person}{Rogelio~E Cardona-Rivera}, \bibinfo{person}{Bradley~A Cassell}, \bibinfo{person}{Stephen~G Ware}, {and} \bibinfo{person}{R~Michael Young}.} \bibinfo{year}{[n.\,d.]}\natexlab{}.
\newblock \showarticletitle{Indexter: A computational model of the event-indexing situation model for characterizing narratives}.
\newblock


\bibitem[Cardona-Rivera and Young(2013)]%
        {cardona2013cognitivist}
\bibfield{author}{\bibinfo{person}{Rogelio~Enrique Cardona-Rivera} {and} \bibinfo{person}{Robert~Michael Young}.} \bibinfo{year}{2013}\natexlab{}.
\newblock \showarticletitle{A Cognitivist Theory of Affordances for Games.}. In \bibinfo{booktitle}{\emph{DiGRA Conference}}.
\newblock


\bibitem[{CD Projekt Red}(2015)]%
        {game:TW3}
\bibfield{author}{\bibinfo{person}{{CD Projekt Red}}.} \bibinfo{year}{2015}\natexlab{}.
\newblock \bibinfo{title}{The Witcher 3: The Wild Hunt}.
\newblock
\newblock
\newblock
\shownote{Video game}.


\bibitem[Compton(2016)]%
        {compton2016so}
\bibfield{author}{\bibinfo{person}{Kate Compton}.} \bibinfo{year}{2016}\natexlab{}.
\newblock \showarticletitle{So you want to build a generator}.
\newblock  (\bibinfo{year}{2016}).
\newblock


\bibitem[{DICE}(2008)]%
        {game:ME}
\bibfield{author}{\bibinfo{person}{{DICE}}.} \bibinfo{year}{2008}\natexlab{}.
\newblock \bibinfo{title}{Mirror's Edge}.
\newblock
\newblock
\newblock
\shownote{Video game}.


\bibitem[Fendt et~al\mbox{.}(2012)]%
        {fendt2012achieving}
\bibfield{author}{\bibinfo{person}{Matthew~William Fendt}, \bibinfo{person}{Brent Harrison}, \bibinfo{person}{Stephen~G Ware}, \bibinfo{person}{Rogelio~E Cardona-Rivera}, {and} \bibinfo{person}{David~L Roberts}.} \bibinfo{year}{2012}\natexlab{}.
\newblock \showarticletitle{Achieving the illusion of agency}. In \bibinfo{booktitle}{\emph{Interactive Storytelling: 5th International Conference, ICIDS 2012, San Sebasti{\'a}n, Spain, November 12-15, 2012. Proceedings 5}}. Springer, \bibinfo{pages}{114--125}.
\newblock


\bibitem[Fern{\'a}ndez-Vara(2011)]%
        {fernandez2011game}
\bibfield{author}{\bibinfo{person}{Clara Fern{\'a}ndez-Vara}.} \bibinfo{year}{2011}\natexlab{}.
\newblock \showarticletitle{Game spaces speak volumes: Indexical storytelling}.
\newblock  (\bibinfo{year}{2011}).
\newblock


\bibitem[{Hello Games Studios}(2016)]%
        {nms}
\bibfield{author}{\bibinfo{person}{{Hello Games Studios}}.} \bibinfo{year}{2016}\natexlab{}.
\newblock \bibinfo{title}{No Man's Sky}.
\newblock
\newblock
\newblock
\shownote{Video game}.


\bibitem[Hendrikx et~al\mbox{.}(2013)]%
        {hendrikx2013procedural}
\bibfield{author}{\bibinfo{person}{Mark Hendrikx}, \bibinfo{person}{Sebastiaan Meijer}, \bibinfo{person}{Joeri Van Der~Velden}, {and} \bibinfo{person}{Alexandru Iosup}.} \bibinfo{year}{2013}\natexlab{}.
\newblock \showarticletitle{Procedural content generation for games: A survey}.
\newblock \bibinfo{journal}{\emph{ACM Transactions on Multimedia Computing, Communications, and Applications (TOMM)}} \bibinfo{volume}{9}, \bibinfo{number}{1} (\bibinfo{year}{2013}), \bibinfo{pages}{1--22}.
\newblock


\bibitem[Herv{\'e} and Salge(2021)]%
        {herve2021comparing}
\bibfield{author}{\bibinfo{person}{Jean-Baptiste Herv{\'e}} {and} \bibinfo{person}{Christoph Salge}.} \bibinfo{year}{2021}\natexlab{}.
\newblock \showarticletitle{Comparing PCG metrics with Human Evaluation in Minecraft Settlement Generation}. In \bibinfo{booktitle}{\emph{Proceedings of the 16th International Conference on the Foundations of Digital Games}}. \bibinfo{pages}{1--15}.
\newblock


\bibitem[Herv{\'e} and Salge(2022)]%
        {herve2022automated}
\bibfield{author}{\bibinfo{person}{Jean-Baptiste Herv{\'e}} {and} \bibinfo{person}{Christoph Salge}.} \bibinfo{year}{2022}\natexlab{}.
\newblock \showarticletitle{Automated Isovist Computation for Minecraft}.
\newblock \bibinfo{journal}{\emph{arXiv preprint arXiv:2204.03752}} (\bibinfo{year}{2022}).
\newblock


\bibitem[Herv{\'e} et~al\mbox{.}(2023a)]%
        {herve2023examination}
\bibfield{author}{\bibinfo{person}{Jean-Baptiste Herv{\'e}}, \bibinfo{person}{Christoph Salge}, {and} \bibinfo{person}{Henrik Warpefelt}.} \bibinfo{year}{2023}\natexlab{a}.
\newblock \showarticletitle{An Examination of the Hidden Judging Criteria in the Generative Design in Minecraft Competition}.
\newblock  (\bibinfo{year}{2023}).
\newblock


\bibitem[Herv{\'e} et~al\mbox{.}(2023b)]%
        {herve2023exploring}
\bibfield{author}{\bibinfo{person}{Jean-Baptiste Herv{\'e}}, \bibinfo{person}{Oliver Withington}, \bibinfo{person}{Marion Herv{\'e}}, \bibinfo{person}{Laurissa Tokarchuk}, {and} \bibinfo{person}{Christoph Salge}.} \bibinfo{year}{2023}\natexlab{b}.
\newblock \showarticletitle{Exploring Minecraft Settlement Generators with Generative Shift Analysis}.
\newblock \bibinfo{journal}{\emph{arXiv preprint arXiv:2309.05371}} (\bibinfo{year}{2023}).
\newblock


\bibitem[Janlert and Stolterman(1997)]%
        {janlert1997character}
\bibfield{author}{\bibinfo{person}{Lars-Erik Janlert} {and} \bibinfo{person}{Erik Stolterman}.} \bibinfo{year}{1997}\natexlab{}.
\newblock \showarticletitle{The character of things}.
\newblock \bibinfo{journal}{\emph{Design Studies}} \bibinfo{volume}{18}, \bibinfo{number}{3} (\bibinfo{year}{1997}), \bibinfo{pages}{297--314}.
\newblock


\bibitem[Khaleque et~al\mbox{.}(2024)]%
        {khaleque2024evaluating}
\bibfield{author}{\bibinfo{person}{Bobby Khaleque}, \bibinfo{person}{Mike Cook}, {and} \bibinfo{person}{Jeremy Gow}.} \bibinfo{year}{2024}\natexlab{}.
\newblock \showarticletitle{Evaluating Environments Using Exploratory Agents}.
\newblock \bibinfo{journal}{\emph{arXiv preprint arXiv:2409.02632}} (\bibinfo{year}{2024}).
\newblock


\bibitem[{Kojima Production}(2019)]%
        {game:DeathStranding}
\bibfield{author}{\bibinfo{person}{{Kojima Production}}.} \bibinfo{year}{2019}\natexlab{}.
\newblock \bibinfo{title}{Death Stranding}.
\newblock
\newblock
\newblock
\shownote{Video game}.


\bibitem[Lee et~al\mbox{.}(2014)]%
        {lee2014facet}
\bibfield{author}{\bibinfo{person}{Jin~Ha Lee}, \bibinfo{person}{Natascha Karlova}, \bibinfo{person}{Rachel~Ivy Clarke}, \bibinfo{person}{Katherine Thornton}, {and} \bibinfo{person}{Andrew Perti}.} \bibinfo{year}{2014}\natexlab{}.
\newblock \showarticletitle{Facet analysis of video game genres}.
\newblock \bibinfo{journal}{\emph{IConference 2014 Proceedings}} (\bibinfo{year}{2014}).
\newblock


\bibitem[Mari{\~n}o et~al\mbox{.}(2015)]%
        {marino2015empirical}
\bibfield{author}{\bibinfo{person}{Julian Mari{\~n}o}, \bibinfo{person}{Willian Reis}, {and} \bibinfo{person}{Levi Lelis}.} \bibinfo{year}{2015}\natexlab{}.
\newblock \showarticletitle{An empirical evaluation of evaluation metrics of procedurally generated Mario levels}. In \bibinfo{booktitle}{\emph{Proceedings of the AAAI Conference on Artificial Intelligence and Interactive Digital Entertainment}}, Vol.~\bibinfo{volume}{11}. \bibinfo{pages}{44--50}.
\newblock


\bibitem[McGrenere and Ho(2000)]%
        {mcgrenere2000affordances}
\bibfield{author}{\bibinfo{person}{Joanna McGrenere} {and} \bibinfo{person}{Wayne Ho}.} \bibinfo{year}{2000}\natexlab{}.
\newblock \showarticletitle{Affordances: Clarifying and evolving a concept}. In \bibinfo{booktitle}{\emph{Graphics interface}}, Vol.~\bibinfo{volume}{2000}. \bibinfo{pages}{179--186}.
\newblock


\bibitem[Mojang(2011)]%
        {game:Minecraft}
\bibfield{author}{\bibinfo{person}{Mojang}.} \bibinfo{year}{2011}\natexlab{}.
\newblock \bibinfo{title}{{M}inecraft}.
\newblock
\newblock
\newblock
\shownote{Video game}.


\bibitem[Sklar(2013)]%
        {sklar2013dream}
\bibfield{author}{\bibinfo{person}{Marty Sklar}.} \bibinfo{year}{2013}\natexlab{}.
\newblock \bibinfo{booktitle}{\emph{Dream It! Do It!: My Half-Century Creating Disney's Magic Kingdoms}}.
\newblock \bibinfo{publisher}{Disney Electronic Content}.
\newblock


\bibitem[Smith and Whitehead(2010)]%
        {smith2010analyzing}
\bibfield{author}{\bibinfo{person}{Gillian Smith} {and} \bibinfo{person}{Jim Whitehead}.} \bibinfo{year}{2010}\natexlab{}.
\newblock \showarticletitle{Analyzing the expressive range of a level generator}. In \bibinfo{booktitle}{\emph{Proceedings of the 2010 workshop on procedural content generation in games}}. \bibinfo{pages}{1--7}.
\newblock


\bibitem[Summerville et~al\mbox{.}(2017)]%
        {summerville2017understanding}
\bibfield{author}{\bibinfo{person}{Adam Summerville}, \bibinfo{person}{Julian~RH Mari{\~n}o}, \bibinfo{person}{Sam Snodgrass}, \bibinfo{person}{Santiago Onta{\~n}{\'o}n}, {and} \bibinfo{person}{Levi~HS Lelis}.} \bibinfo{year}{2017}\natexlab{}.
\newblock \showarticletitle{Understanding mario: an evaluation of design metrics for platformers}. In \bibinfo{booktitle}{\emph{Proceedings of the 12th international conference on the foundations of digital games}}. \bibinfo{pages}{1--10}.
\newblock


\bibitem[Summerville et~al\mbox{.}(2018)]%
        {summerville2018procedural}
\bibfield{author}{\bibinfo{person}{Adam Summerville}, \bibinfo{person}{Sam Snodgrass}, \bibinfo{person}{Matthew Guzdial}, \bibinfo{person}{Christoffer Holmg{\aa}rd}, \bibinfo{person}{Amy~K Hoover}, \bibinfo{person}{Aaron Isaksen}, \bibinfo{person}{Andy Nealen}, {and} \bibinfo{person}{Julian Togelius}.} \bibinfo{year}{2018}\natexlab{}.
\newblock \showarticletitle{Procedural content generation via machine learning (PCGML)}.
\newblock \bibinfo{journal}{\emph{IEEE Transactions on Games}} \bibinfo{volume}{10}, \bibinfo{number}{3} (\bibinfo{year}{2018}), \bibinfo{pages}{257--270}.
\newblock


\bibitem[{Togelius} et~al\mbox{.}(2010)]%
        {Baumgarten2010}
\bibfield{author}{\bibinfo{person}{J. {Togelius}}, \bibinfo{person}{S. {Karakovskiy}}, {and} \bibinfo{person}{R. {Baumgarten}}.} \bibinfo{year}{2010}\natexlab{}.
\newblock \showarticletitle{The 2009 Mario AI Competition}. In \bibinfo{booktitle}{\emph{IEEE Congress on Evolutionary Computation}}. \bibinfo{pages}{1--8}.
\newblock
\urldef\tempurl%
\url{https://doi.org/10.1109/CEC.2010.5586133}
\showDOI{\tempurl}


\bibitem[{Togelius} et~al\mbox{.}(2011)]%
        {Togelius2011}
\bibfield{author}{\bibinfo{person}{J. {Togelius}}, \bibinfo{person}{G.~N. {Yannakakis}}, \bibinfo{person}{K.~O. {Stanley}}, {and} \bibinfo{person}{C. {Browne}}.} \bibinfo{year}{2011}\natexlab{}.
\newblock \showarticletitle{Search-Based Procedural Content Generation: A Taxonomy and Survey}.
\newblock \bibinfo{journal}{\emph{IEEE Transactions on Computational Intelligence and AI in Games}} \bibinfo{volume}{3}, \bibinfo{number}{3} (\bibinfo{year}{2011}), \bibinfo{pages}{172--186}.
\newblock
\urldef\tempurl%
\url{https://doi.org/10.1109/TCIAIG.2011.2148116}
\showDOI{\tempurl}


\bibitem[{Unknown Worlds Entertainment}(2018)]%
        {game:Subnautica}
\bibfield{author}{\bibinfo{person}{{Unknown Worlds Entertainment}}.} \bibinfo{year}{2018}\natexlab{}.
\newblock \bibinfo{title}{Subnautica}.
\newblock
\newblock
\newblock
\shownote{Video game}.


\bibitem[Valls-Vargas et~al\mbox{.}(2015)]%
        {valls2015exploring}
\bibfield{author}{\bibinfo{person}{Josep Valls-Vargas}, \bibinfo{person}{Santiago Ontan{\'o}n}, {and} \bibinfo{person}{Jichen Zhu}.} \bibinfo{year}{2015}\natexlab{}.
\newblock \showarticletitle{Exploring player trace segmentation for dynamic play style prediction}. In \bibinfo{booktitle}{\emph{Proceedings of the AAAI conference on artificial intelligence and interactive digital entertainment}}, Vol.~\bibinfo{volume}{11}. \bibinfo{pages}{93--99}.
\newblock


\bibitem[Warpefelt(2020)]%
        {warpefelt2020micro}
\bibfield{author}{\bibinfo{person}{Henrik Warpefelt}.} \bibinfo{year}{2020}\natexlab{}.
\newblock \showarticletitle{Micro-level examination of games using {Indicator} {Analysis}}. In \bibinfo{booktitle}{\emph{{FDG} '20: {International} {Conference} on the {Foundations} of {Digital} {Games}}}. \bibinfo{publisher}{ACM}, \bibinfo{pages}{1--9}.
\newblock
\urldef\tempurl%
\url{https://doi.org/10.1145/3402942.3402980}
\showDOI{\tempurl}


\bibitem[Yannakakis et~al\mbox{.}(2013)]%
        {yannakakis2013player}
\bibfield{author}{\bibinfo{person}{Georgios~N Yannakakis}, \bibinfo{person}{Pieter Spronck}, \bibinfo{person}{Daniele Loiacono}, {and} \bibinfo{person}{Elisabeth Andr{\'e}}.} \bibinfo{year}{2013}\natexlab{}.
\newblock \showarticletitle{Player modeling}.
\newblock  (\bibinfo{year}{2013}).
\newblock


\bibitem[Yannakakis and Togelius(2011)]%
        {yannakakis2011experience}
\bibfield{author}{\bibinfo{person}{Georgios~N Yannakakis} {and} \bibinfo{person}{Julian Togelius}.} \bibinfo{year}{2011}\natexlab{}.
\newblock \showarticletitle{Experience-driven procedural content generation}.
\newblock \bibinfo{journal}{\emph{IEEE Transactions on Affective Computing}} \bibinfo{volume}{2}, \bibinfo{number}{3} (\bibinfo{year}{2011}), \bibinfo{pages}{147--161}.
\newblock


\end{thebibliography}

\appendix

\end{document}